# An Extended Kalman Filter Integrated Latent Feature Model on Dynamic Weighted Directed Graphs

Hongxun Zhou, Xiangyu Chen, and Ye Yuan

*Abstract*— **A dynamic weighted directed graph (DWDG) is commonly encountered in various application scenarios. It involves extensive dynamic interactions among numerous nodes. Most existing approaches explore the intricate temporal patterns hidden in a DWDG from the purely data-driven perspective, which suffers from accuracy loss when a DWDG exhibits strong fluctuations over time. To address this issue, this study proposes a novel Extended-Kalman-Filter-Incorporated Latent Feature (EKLF) model to represent a DWDG from the model-driven perspective. Its main idea is divided into the following two-fold ideas: a) adopting a control model, i.e., the Extended Kalman Filter (EKF), to track the complex temporal patterns precisely with its nonlinear state-transition and observation functions; and b) introducing an alternating least squares (ALS) algorithm to train the latent features (LFs) alternatively for precisely representing a DWDG. Empirical studies on DWDG datasets demonstrate that the proposed EKLF model outperforms state-of-the-art models in prediction accuracy and computational efficiency for missing edge weights of a DWDG. It unveils the potential for precisely representing a DWDG by incorporating a control model.**

## I. Introduction

A dynamic weighted directed graph (DWDG) is a mathematical abstraction that represents a system of entities and the evolving relationships between them over time [1-4]. Specifically, it incorporates directionality in relationships and assigns weights to the edges to capture the strength or significance of those relationships. Such a DWDG contains a great deal of valuable knowledge and is often encountered in various real scenarios, i.e., transportation networks [1], social networks [2], and communication networks [3]. Hence, how to perform a precise representation of a DWDG for various data analysis tasks has become a highly popular issue [1-6].

In recent years, Graph Neural Networks (GNNs) have gained popularity since they provide a powerful framework for analyzing and modeling graph-structured data based on the message-passing mechanism. For instance, He et al. [7] propose the lightGCN, which eliminates the feature transformation and nonlinear activation of the traditional graph convolution network (GCN). Yun et al. [8] propose GTN which can represent heterogeneous information in the network and identify unconnected but potentially useful edges in the original graph. Guo et al. [9] propose DGCN-HN which adds residual connection and global connection. However, the previous models are static models, which cannot capture a DWDG accurately. To address it, a pyramid of sophisticated dynamic GNN models is proposed. For instance, Manessi et al. [10] propose WD-GCN which is a combination of graph convolution units and LSTM. Malik et al. [11] propose TM-GCN which extends the GCN to learn representations of dynamic graphs using the m-product technique.

It is pointed out that a DWDG can be denoted by an incomplete matrix sequence [12-17], as shown in Fig. 1, since it becomes impossible to observe the full interactions among nodes at each time slot due to uncontrollable reasons, i.e., sensor fault in transportation networks [1, 22-24]. According to existing studies, a latent feature (LF) model [18-21, 25-27, 31-33] is also a mainstream approach to perform the representation of a DWDG since it proves to be highly effective in addressing the incomplete matrix sequence. For instance, Luo et al. [29] present a biased tensor-based LF model, which incorporates linear biases to represent the temporal patterns accurately. Zhang et al. [28] propose a non-negative tensor-based LF model to handle an incomplete matrix sequence while considering the non-negative constraints. Wu et al. [30] propose diversified tensor-based LF models to extract desired features from the target DWDG.

Although the above models obtain convincing performance on a DWDG, they explore the complex temporal patterns hidden in a DWDG from the purely data-driven perspective, which suffers from accuracy loss when a DWDG exhibits strong fluctuations over time [37]. Fig. 2 gives an illustrative example of such time-dependent data fluctuations.

Hence, it appears essential to implement a tailored model design to characterize such fluctuations in a DWDG. As shown in [34, 36], an Extended Kalman Filter is a recursive control model designed for a nonlinear system in control theory, which is commonly employed to handle various problems involving dynamic estimation with uncertain fluctuations [35, 38, 39] It can estimate each temporal state of a dynamic system based on previous ones, precisely representing temporal patterns with its nonlinear state-transition and observation functions.

Motivated by this discovery, this study proposes a novel Extended-Kalman-Filter-Incorporated Latent Feature (EKLF) model with the following two-fold ideas:

➢ H. X. Zhou and Y. Yuan are with the College of Computer and Information Science, Southwest University, Chongqing 400715, China (e-mail: libnova7@swu.edu.cn, yuanyekl@swu.edu.cn).
➢ X. Y. Chen is with the Hanhong College, Southwest University, Chongqing 400715, China (e-mail: chenrt213@outlook.com).

- Adopting a control model, i.e., the Extended Kalman Filter (EKF), to track the intricate temporal patterns precisely with its nonlinear state-transition and observation functions;
- Introducing an alternating least squares (ALS) algorithm to train the LFs alternatively for precisely representing a DWDG.

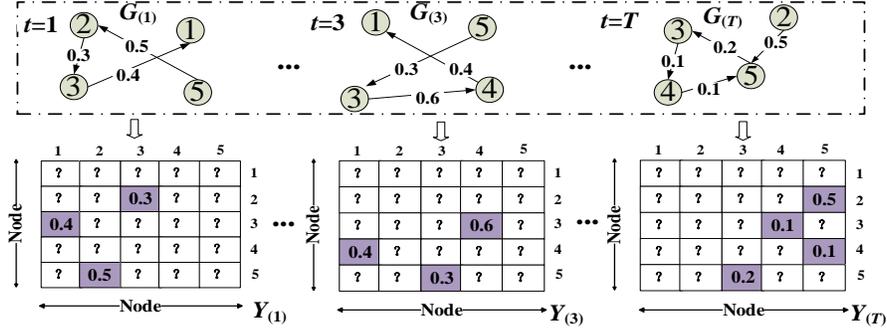

Figure 1. A DWDG and its corresponding incomplete matrix sequence.

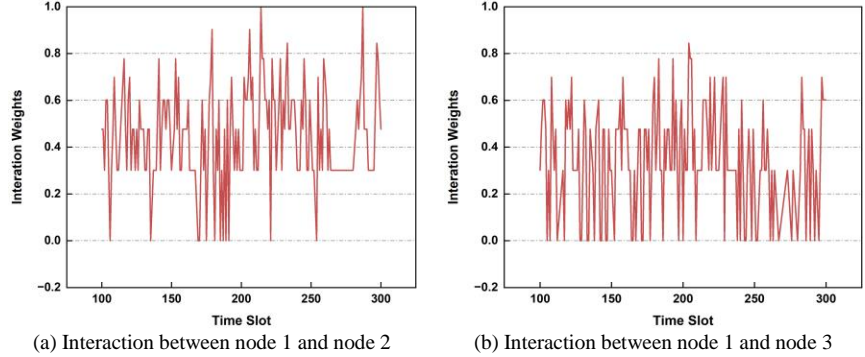

(a) Interaction between node 1 and node 2  (b) Interaction between node 1 and node 3

Figure 2. Time-dependent data fluctuations of a DWDG.

The main contribution of this study is that the proposed EKLF model effortlessly integrates the principle of an EKF into its learning framework to represent a DWDG [40-42, 82] from the model-driven perspective. Specifically, it can precisely capture the strong data fluctuations with the nonlinear state-transition and observation functions of an EKF. It provides a direction for precisely representing a DWDG by incorporating a control model in control theory.

Empirical studies on DWDG datasets demonstrate that the proposed EKLF model outperforms state-of-the-art models in prediction accuracy and computational efficiency for missing edge weights of a DWDG.

## II. PRELIMINARIES

### A. Problem Formulation

Note that a Dynamic Weighted Directed Graph and an Incomplete Matrix Sequence are defined as follows:

***Definition* 1 (Dynamic Weighted Directed Graph)**: Note that this study primarily focuses on the changes in the edges over time. Hence, a DWDG is defined as $G=\{G_{(1)}, G_{(2)}, …, G_{(T)}\}$ and each graph snapshot $G_{(t)} = (V, E_{(t)})$ is a graph at time slot $t$ ($1 \leq t \leq T$), where $V$ is the node set with $M$ nodes, and $E_{(t)}$ indicates the edge set at time $t$ between node $i$ and node $j$.

***Definition* 2 (Incomplete Matrix Sequence)**: Naturally, each graph snapshot $G_{(t)} = (V, E_{(t)})$ in a DWDG can be described by an adjacency matrix $Y_{(t)}$, as shown in Fig. 1. In addition, a quantifiable index is allocated to the edge between two nodes as its weight. Specifically, according to Fig. 1, only a few directed edges are observed and the others are missing. Hence, $Y_{(t)}$ is an incomplete matrix. Ultimately, a DWDG $G=\{G_{(1)}, G_{(2)}, …, G_{(T)}\}$ can be reformulated by an incomplete matrix sequence $Y=\{Y_{(1)}, Y_{(2)}, …, Y_{(T)}\}$.

Hence, in our context, performing a representation of a DWDG is equivalent to analyzing its corresponding incomplete matrix sequence.

### B. A Latent Feature Model

A latent feature (LF) model [43-51, 76-81], is widely adopted for addressing an incomplete matrix sequence. Given a target incomplete matrix sequence $Y=\{Y_{(1)}, Y_{(2)}, …, Y_{(T)}\}$, it builds the rank-$f$ approximation to each $Y_{(t)}$ by two LFs $N_{(t)}$ and $Q_{(t)}$ on $Y_{(t)}$'s known entity set $Y_{(t)\Lambda}$. To achieve this goal, an objective function on the regularized Euclidean distance is defined as:

$$\varepsilon(Y) = \varepsilon(N,Q) = \sum_{t=1}^{T} \sum_{y_{(t)i,j} \in Y_{(t)\Lambda}} \begin{pmatrix} \lambda \left( y_{(t)i,j} - \left\langle n_{(t)i}, q_{(t)j} \right\rangle \right)^2 \\ + \left\| n_{(t)i} \right\|_2^2 + \left\| q_{(t)j} \right\|_2^2 \end{pmatrix} \quad (1)$$

where $y_{(t)i,j}$ is the single element of $Y_{(t)}$, $\lambda$ is regularization coefficient, $<,>$ computes the inner product of two compatible vectors, $\|\cdot\|_2$ is $L_2$ norm to prevent overfitting, $n_{(t)i}$ and $q_{(t)j}$ denote the $i$-th row vector and $j$-th row vector of $N_{(t)}$ and $Q_{(t)}$, respectively.

## III. METHOD

For performing precise representation of a DWDG, the proposed EKLF contains the following two procedures:
- **An EKF-based *N*-procedure**. Utilizing an EKF to track the intricate temporal patterns accurately, thereby obtaining the temporal LFs $N=\{N_{(1)}, N_{(2)}, \ldots, N_{(T)}\}$;
- **An ALS-based *Q*-procedure.** Implementing an ALS algorithm for finely representing the historical data, thereby achieving the time-consistent LFs $Q=\{Q_{(1)}, Q_{(2)}, \ldots, Q_{(T)}\}$.

### A. An EKF-based N-procedure

To capture the intricate temporal patterns within a DWDG, we initially construct a coupled dynamic system involving temporal LFs $N=\{N_{(1)}, N_{(2)}, \ldots, N_{(T)}\}$, whose posterior state can be estimated by the EKF. Specifically, such a coupled dynamic system is described by a state-transition process and an observation process.

To describe the dynamic variation of temporal LFs $N=\{N_{(1)}, N_{(2)}, \ldots, N_{(T)}\}$, the nonlinear state-transition function $S(\cdot)$ and the state-transition noise $w_{(t)i}$ are adopted to achieve the state process equation:

$$n_{(t)i} = S\left(n_{(t-1)i}\right) + w_{(t)i} \quad (2)$$

where $S(\cdot)$ denotes a nonlinear state-transition function, $n_{(t)i}$ denotes the state vectors of node $i$ at time $t$, and $w_{(t)i}$ is state-transition noise with $w_{(t)i} \sim N(0, W_{(t)i})$ and $W_{(t)i}$ is its covariance matrix. Further, we expand $S(n_{(t-1)i})$ for the first-order Taylor's series at $\hat{n}^+_{(t-1)i}$ as follows:

$$\begin{aligned} S\left(n_{(t-1)i}\right) &\approx S\left(\hat{n}^+_{(t-1)i}\right) + S'\left(\hat{n}^+_{(t-1)i}\right)\left(n_{(t-1)i} - \hat{n}^+_{(t-1)i}\right) \\ &\approx S'\left(\hat{n}^+_{(t-1)i}\right) n_{(t-1)i} + \left(S\left(\hat{n}^+_{(t-1)i}\right) - S'\left(\hat{n}^+_{(t-1)i}\right)\hat{n}^+_{(t-1)i}\right) \\ &\approx B_{(t-1)i} n_{(t-1)i} + C_{(t-1)i} \end{aligned} \quad (3)$$

where $\hat{n}^+_{(t-1)i}$ is the posterior state of $n_{(t-1)i}$, $B_{(t-1)i} = S'\left(\hat{n}^+_{(t-1)i}\right)$ and $C_{(t-1)i} = S\left(\hat{n}^+_{(t-1)i}\right) - S'\left(\hat{n}^+_{(t-1)i}\right)\hat{n}^+_{(t-1)i}$. On the other hand, following the principle of an EKF, the following observation process equation is built:

$$y_{(t)i} = O\left(n_{(t)i}\right) + r_{(t)i} \quad (4)$$

where $y_{(t)i}$ denotes observation edges related to node $i$ at time slot $t$, $O(\cdot)$ denotes a nonlinear observation function, and $r_{(t)i}$ denotes observation noise with $r_{(t)i} \sim N(0, R_{(t)i})$ and $R_{(t)i}$ is its covariance matrix. Further, we expand $O(n_{(t)i})$ for the first-order Taylor's series at $\hat{n}^-_{(t)i}$ as follows:

$$\begin{aligned} O\left(n_{(t)i}\right) &\approx O\left(\hat{n}^-_{(t)i}\right) + O'\left(\hat{n}^-_{(t)i}\right)\left(n_{(t)i} - \hat{n}^-_{(t)i}\right) \\ &\approx O'\left(\hat{n}^-_{(t)i}\right) n_{(t)i} + \left(O\left(\hat{n}^-_{(t)i}\right) - O'\left(\hat{n}^-_{(t)i}\right)\hat{n}^-_{(t)i}\right) \\ &\approx D_{(t)i} n_{(t)i} + H_{(t)i} \end{aligned} \quad (5)$$

where $\hat{n}^-_{(t)i}$ is the prior state of $n_{(t)i}$, $D_{(t)i} = O'\left(\hat{n}^-_{(t)i}\right)$ and $H_{(t)i} = O\left(\hat{n}^-_{(t)i}\right) - O'\left(\hat{n}^-_{(t)i}\right)\hat{n}^-_{(t)i}$.

Utilizing the deductions above, we construct a coupled dynamic system following the EKF, which adopts two steps to obtain the desired temporal LFs:

**1) State Prediction:**

$$B_{(t-1)i} = S'\left(\hat{n}^+_{(t-1)i}\right) \quad (6)$$

$$\hat{n}^-_{(t)i} = S\left(\hat{n}^+_{(t-1)i}\right) \quad (7)$$

$$\hat{P}^-_{(t)i} = B_{(t-1)i} \hat{P}^+_{(t-1)i} B^T_{(t-1)} + W_{(t)i} \quad (8)$$

where $\hat{n}^-_{(t)i}$ denotes a prior estimate of $n_{(t)i}$, $\hat{P}^-_{(t)i}$ denotes the prior covariance matrix of $n_{(t)i}$. Following the principle of EKF, we revise the prior information $\hat{n}^-_{(t)i}$ and $\hat{P}^-_{(t)i}$ to obtain the final posterior state of $n_{(t)i}$ with the following step:

**2) Feedback update:**

$$D_{(t)i} = O'\left(\hat{n}^-_{(t)i}\right) = Q_{(t)i} f'\left(\hat{n}^-_{(t)i}\right) \quad (9)$$

$$K_{(t)i} = \frac{\hat{P}_{(t)i}^{-} D_{(t)i}^{T}}{D_{(t)i}\hat{P}_{(t)i}^{-} D_{(t)i}^{T} + R_{(t)i}} \qquad (10)$$

$$\hat{n}_{(t)i}^{+} = \hat{n}_{(t)i}^{-} + K_{(t)i}\left(y_{(t)i} - O\left(\hat{n}_{(t)i}^{-}\right)\right) \qquad (11)$$

$$\hat{P}_{(t)i}^{+} = \hat{P}_{(t)i}^{-} - K_{(t)i} D_{(t)i} \hat{P}_{(t)i}^{-} \qquad (12)$$

where $K_{(t)i}$ is an extended Kalman gain, $\hat{n}_{(t)i}^{+}$ denotes a posterior estimate of $n_{(t)i}$, $Q_{(t)i}$ is a collection of $q_{(t)i}$ invoked by node $i$ at time slot $t$, $\hat{P}_{(t)i}^{+}$ denotes the posterior covariance matrix of $n_{(t)i}$. With (6-12), $\forall t \in T$ and $i \in M$, the temporal LFs $N=\{N_{(1)}, N_{(2)}, \ldots, N_{(T)}\}$ is obtained.

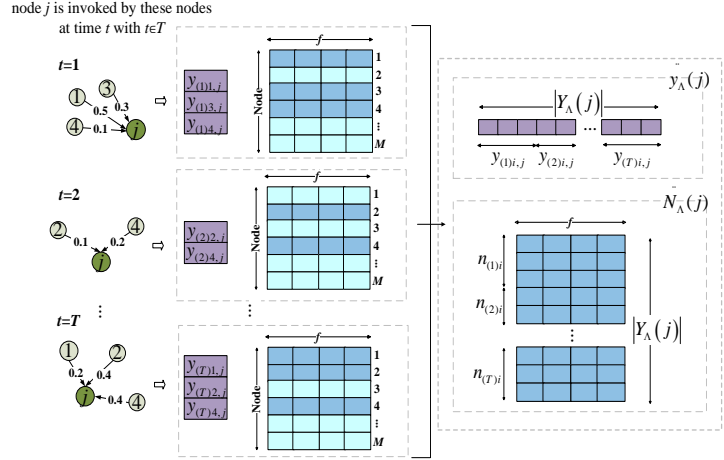

Figure 3. The building process of $\ddot{y}_\Lambda(j)$ and $\ddot{N}_\Lambda(j)$.

## B. An ALS-based Q-procedure

Particularly, the temporal LFs $N=\{N_{(1)}, N_{(2)}, \ldots, N_{(T)}\}$ is obtained based on EKF to describe the intricate temporal patterns. Hence, in an ALS-based Q-procedure, we consider $Q$ to be time-consistent, i.e., $Q_{(1)}=Q_{(2)}=, \ldots, =Q_{(T)}$, for easy convenient calculation. Under this assumption, the following objective function with $Q$ is achieved by simplifying objective function (1):

$$\varepsilon(Y) = \varepsilon(N,Q) = \sum_{t=1}^{T}\sum_{y_{(t)i,j} \in Y_{(t)\Lambda}} \left( \begin{array}{c} \lambda\left(y_{(t)i,j} - \langle n_{(t)i}, q_j \rangle\right)^2 \\ + \|n_{(t)i}\|_2^2 + \|q_j\|_2^2 \end{array} \right). \qquad (13)$$

Naturally, the partial loss $\varepsilon(q_j)$ with $j \in M$ is given as:

$$\varepsilon(q_j) = \sum_{t=1}^{T}\sum_{y_{(t)i,j} \in Y_{(t)\Lambda}(j)} \left( \lambda\left(y_{(t)i,j} - \langle n_{(t)i}, q_j \rangle\right)^2 + \|q_j\|_2^2 \right) \qquad (14)$$

where $Y_{(t)\Lambda}(j)$ is the $Y_{(t)\Lambda}$ subset related to the node $j$. Note that we can get the following linear equation system by (14):

$$\frac{\delta\varepsilon(q_j)}{\delta q_j} = \begin{cases} \forall i \in Y_{(1)\Lambda(j)} : 2\lambda\left(y_{(1)i,j} - n_{(1)i}\cdot q_j\right)\left(-n_{(1)i}\right) + 2q_j \\ \vdots \\ \forall i \in Y_{(T)\Lambda(j)} : 2\lambda\left(y_{(T)i,j} - n_{(T)i}\cdot q_j\right)\left(-n_{(T)i}\right) + 2q_j \end{cases}$$

$$\Rightarrow \frac{\delta\varepsilon(q_j)}{\delta q_j} = part1 + part2 + part3 = 0 \qquad (15)$$

where

$$\begin{aligned} part1 &= -2\lambda\, \ddot{y}_\Lambda(j)\ddot{N}_\Lambda(j), \\ part2 &= 2\lambda q_j \left(\ddot{N}_\Lambda(j)\right)^T \ddot{N}_\Lambda(j), \\ part3 &= 2|Y_\Lambda(j)|q_j. \end{aligned} \qquad (16)$$

The building process of $\ddot{y}_\Lambda(j)$ and $\ddot{N}_\Lambda(j)$ is given in Fig. 3. Based on (15) and (16), $q_j$ is achieved as follows:

$$q_j = \frac{\ddot{y}_\Lambda(j)\ddot{N}_\Lambda(j)}{\left(\ddot{N}_\Lambda(j)\right)^T \ddot{N}_\Lambda(j) + \frac{|Y_\Lambda(j)|I}{\lambda}} \tag{17}$$

where *I* denotes the identity matrix. By applying (17) to all $q_j$ as $\forall j \in J$, the learning rule of time-consistent LFs *Q* is achieved based on an ALS algorithm.

Then by iteratively executing the EKF-based *N*-procedure and ALS-based *Q*-procedure until achieving a stationary solution, we finally obtain an EKLF model with considering the temporal dynamics hidden in a DWDG. In our context, we adopt the missing edge weight estimation of a DWDG as the downstream task. Hence, we obtain the estimate $\tilde{Y}_{(t)}$ as follows:

$$\tilde{Y}_{(t)} = N_{(t)}Q^T. \tag{18}$$

Based on the above design, each iteration of EKLF contains two procedures: a) executing the EKF-based *N*-procedure to obtain the temporal LFs $N=\{N_{(1)}, N_{(2)}, \ldots, N_{(T)}\}$ to reveal the temporal patterns hidden in a DWDG; b) executing the ALS-based *Q*-procedure to solve the time-consistent LFs *Q*.

IV. EMPIRICAL STUDIES

*A. General Settings*

**Evaluation Protocol:** This algorithm aims to estimate the missing edge weight estimation of a DWDG. Commonly, root mean square error (RMSE) and mean absolute error (MAE) are adopted to evaluate the estimation accuracy [59-62]. The programming is JAVA SE. Empirical studies are conducted on a Tablet with Intel(R) Xeon(R) Gold 5218 CPU @ 2.30GHz, and 512GB RAM.

**Datasets:** We adopt three real DWDG datasets in our empirical analysis part, which are collected from a real terminal interaction pattern analysis system [35, 38]. The detailed information on datasets is recorded in Table I. For each dataset, we design three testing cases, as shown in Table II. The termination stipulation is, i.e., the error threshold is $10^{-5}$, the iteration threshold is 500, and the training process stops if any threshold is satisfied.

**Model Setting:** The basic setting of this experiment is as follows:
1) For the proposed EKLF, 'LeakyReLU' is used in the nonlinear state-transition function and nonlinear observation function;
2) The dimension of the node LF is 20 for all the compared models;
3) we apply a grid search for the compared models with the learning rate $\eta=[0.0001, 0.01]$ and regularization coefficient $\lambda=[0.001, 0.1]$ to achieve their optimal results.

*B. Comparison*

We commence our empirical investigations by comparing the proposed EKLF model with the state-of-the-art models in terms of estimation accuracy and computational efficiency. Tables III and IV show the comparison results on estimation errors and computational efficiency. The model used in this experiment is as follows:

TABLE I. EXPERIMENTAL DATASET DETAILS

| No. | Nodes | Time slots | Known Entries | Density |
|---|---|---|---|---|
| D1 | 499 | 1148 | 29632 | 0.0104% |
| D2 | 1894 | 149 | 29528 | 0.0055% |
| D3 | 1999 | 149 | 30802 | 0.0052% |

TABLE II. TESTING CASES

| No. | Cases | Train-Validation-Test |
|---|---|---|
| D1 | D11 | 10%-10%-80% |
|  | D12 | 20%-10%-70% |
|  | D13 | 30%-10%-60% |
| D2 | D21 | 10%-10%-80% |
|  | D22 | 20%-10%-70% |
|  | D23 | 30%-10%-60% |
| D3 | D31 | 10%-10%-80% |
|  | D32 | 20%-10%-70% |
|  | D33 | 30%-10%-60% |

- **M1 (EKLF):** The proposed model of this study
- **M2 (TeDCaN [63]):** A dynamic model that adopts the graph Laplacian regularization based on tensor decomposition.
- **M3 (TM-GCN [11])**: A dynamic model which extends the GCN to learn representations of dynamic graphs using the tensor m-product.
- **M4 (WD-GCN [10])**: A dynamic model which combines graph convolution units and LSTM.

- **M5 (LightGCNr-AdjNorm [64])**: A static lightGCN model which proposes r-AdjNorm strategy to balance the accuracy and novelty.
- **M6 (DGCN-HN [9])**: A static GCN model which adds residual connection and global connection.
- **M7 (SGL-ED [65])**: A static GCN model which adds auxiliary self-supervised tasks to reinforce node representation learning.
- **M8 (GDE [66]):** A static model based on a graph denoising encoder, which can alleviate the over-smoothing problem.

TABLE III. THE COMPARISON RESULTS ON RMSE/MAE

| Case | | M1 | M2 | M3 | M4 | M5 | M6 | M7 | M8 |
|---|---|---|---|---|---|---|---|---|---|
| D11 | RMSE | **0.2719** | 0.4011 | 0.4091 | 0.4425 | 0.5209 | 0.5178 | 0.5217 | 0.5205 |
| | MAE | **0.1808** | 0.3356 | 0.3155 | 0.3206 | 0.3720 | 0.3722 | 0.3727 | 0.3722 |
| D12 | RMSE | **0.2620** | 0.3692 | 0.4038 | 0.4148 | 0.5119 | 0.5080 | 0.5136 | 0.5117 |
| | MAE | **0.1735** | 0.3058 | 0.3210 | 0.3143 | 0.3643 | 0.3653 | 0.3665 | 0.3658 |
| D13 | RMSE | **0.2565** | 0.3576 | 0.4001 | 0.4532 | 0.5038 | 0.4990 | 0.5054 | 0.5027 |
| | MAE | **0.1698** | 0.2885 | 0.3194 | 0.3186 | 0.3605 | 0.3616 | 0.3622 | 0.3616 |
| D21 | RMSE | **0.2937** | 0.3694 | 0.3821 | 0.3858 | 0.5036 | 0.4907 | 0.5005 | 0.5033 |
| | MAE | **0.1924** | 0.2792 | 0.2902 | 0.2955 | 0.3483 | 0.3470 | 0.3503 | 0.3487 |
| D22 | RMSE | **0.2883** | 0.3683 | 0.3773 | 0.3821 | 0.5008 | 0.4767 | 0.4927 | 0.5006 |
| | MAE | **0.1830** | 0.2767 | 0.2844 | 0.2833 | 0.3459 | 0.3508 | 0.3479 | 0.3472 |
| D23 | RMSE | **0.2725** | 0.3694 | 0.3882 | 0.3994 | 0.4982 | 0.4629 | 0.4866 | 0.4981 |
| | MAE | **0.1768** | 0.2806 | 0.2921 | 0.2972 | 0.3440 | 0.3339 | 0.3474 | 0.3458 |
| D31 | RMSE | **0.3054** | 0.3998 | 0.3827 | 0.3928 | 0.5050 | 0.4911 | 0.5064 | 0.5045 |
| | MAE | **0.1956** | 0.3261 | 0.2927 | 0.2943 | 0.3496 | 0.3469 | 0.3494 | 0.3498 |
| D32 | RMSE | **0.2813** | 0.3741 | 0.3807 | 0.4172 | 0.4999 | 0.4776 | 0.4918 | 0.5001 |
| | MAE | **0.1827** | 0.2942 | 0.2844 | 0.3009 | 0.3451 | 0.3398 | 0.3466 | 0.3462 |
| D33 | RMSE | **0.2714** | 0.3694 | 0.3834 | 0.3861 | 0.4915 | 0.4614 | 0.4791 | 0.4917 |
| | MAE | **0.1770** | 0.2898 | 0.2855 | 0.2873 | 0.3381 | 0.3321 | 0.3408 | 0.3399 |

TABLE IV. THE COMPARISON RESULTS ON TOTAL TIME COST (SECS)

| Cases | | M1 | M2 | M3 | M4 | M5 | M6 | M7 | M8 |
|---|---|---|---|---|---|---|---|---|---|
| D11 | Time-RMSE | **14.4** | 10363.4 | 1300.3 | 34070.6 | 120.1 | 188.9 | 202.3 | 229.3 |
| | Time-MAE | **13.0** | 10363.3 | 1280.3 | 34068.6 | 120.4 | 188.3 | 202.1 | 229.2 |
| D12 | Time-RMSE | **6.0** | 15164.8 | 751.63 | 34446.1 | 134.9 | 177.1 | 304.6 | 278.4 |
| | Time-MAE | **5.2** | 15162.2 | 727.2 | 35881.4 | 134.5 | 177.6 | 304.4 | 278.3 |
| D13 | Time-RMSE | **8.4** | 15142.4 | 929.4 | 32894.4 | 136.2 | 184.3 | 353.1 | 297.9 |
| | Time-MAE | **8.6** | 15140.4 | 787.5 | 28289.2 | 136.9 | 184.9 | 353.7 | 297.6 |
| D21 | Time-RMSE | 21.1 | 38940.2 | 4323.3 | 2374.7 | **15.4** | 20.7 | 60.3 | 244.2 |
| | Time-MAE | **14.4** | 38938.6 | 4320.1 | 2310.5 | 15.2 | 20.4 | 60.2 | 245.8 |
| D22 | Time-RMSE | 15.4 | 36332.6 | 975.7 | 2686.1 | **15.3** | 21.6 | 63.9 | 240.5 |
| | Time-MAE | **12.8** | 34879.8 | 957.3 | 2558.9 | 15.2 | 21.1 | 63.7 | 240.9 |
| D23 | Time-RMSE | **12.2** | 36332.2 | 363.2 | 2901.7 | 14.5 | 20.3 | 61.5 | 238.3 |
| | Time-MAE | **12.2** | 36330.1 | 361.2 | 2379.4 | 14.1 | 20.7 | 61.5 | 237.2 |
| D31 | Time-RMSE | **15.4** | 18840.9 | 202.2 | 3420.4 | 16.3 | 22.5 | 68.4 | 270.8 |
| | Time-MAE | 19.8 | 18837.2 | 280.1 | 3302.4 | **16. 3** | 22.4 | 68.6 | 269.5 |
| D32 | Time-RMSE | 16.0 | 41740.9 | 177.1 | 5549.3 | **15.7** | 21.4 | 69.1 | 278.1 |
| | Time-MAE | **14.8** | 41738.4 | 180.2 | 5438.5 | 15.5 | 21.8 | 69.7 | 277.9 |
| D33 | Time-RMSE | 20.4 | 52936.1 | 194.6 | 5371.8 | **16.5** | 23.4 | 72.9 | 267.5 |
| | Time-MAE | 17.0 | 52936.9 | 177.1 | 5264.3 | **16.7** | 23.7 | 73.6 | 267.6 |

From them, we achieve some important verdicts:
a) **EKLF exhibits a notable advantage in capturing the representation of a DWDG**. As shown in Table III and Figs. 4 - 5, we evidently see that the proposed EKLF outperforms its peers on all the testing cases on the estimation accuracy. For instance, on D13, EKLF obtains the lowest RMSE at 0.2565, which is 28.27% lower than 0.3576 obtained by M2, 35.89% lower than 0.4001 obtained by M3, 43.40% lower than 0.4532 obtained by M4, 49.09% lower than 0.5038 obtained by M5, 48.60% lower than 0.4990 obtained by M6, 49.25% lower than 0.5054 obtained by M7, and 48.98% lower than 0.5027 obtained by M8. Note that the dynamic models, i.e., M1-M4, are significantly outperformed by the static models, i.e., M5-M8. The main reason is that the static models do not consider the temporal patterns hidden in a DWDG. Specifically, compared to the dynamic models, i.e., M2-M4, M1's estimating accuracy is higher. This phenomenon suggests that M1 excels at representing intricate temporal patterns concealed within a DWDG;
b) **EKLF obtains competitive computational efficiency when addressing a DWDG.** As shown in Table IV, M1 achieves the lowest total time cost on 12 cases out of 18 in total. For example, on D11, M1 takes 14.4 seconds to converge in RMSE, which is about 99.86% lower than M2's 10363.4 seconds, 99.89% lower than M3's 1300.3 seconds, 99.96% lower than M4's 34070.6 seconds, 88.01% lower than M5's 120.1 seconds, 92.37% lower than M6's 188.9 seconds, 92.88% lower than M7's 202.3 seconds, and 93.72% lower than M8's 229.3 seconds. The main reason is that the proposed EKLF does not require complex graph convolution operations. In some cases, the static model M5 may compute faster than M1 due to the sparsity and fewer time slices of these datasets.

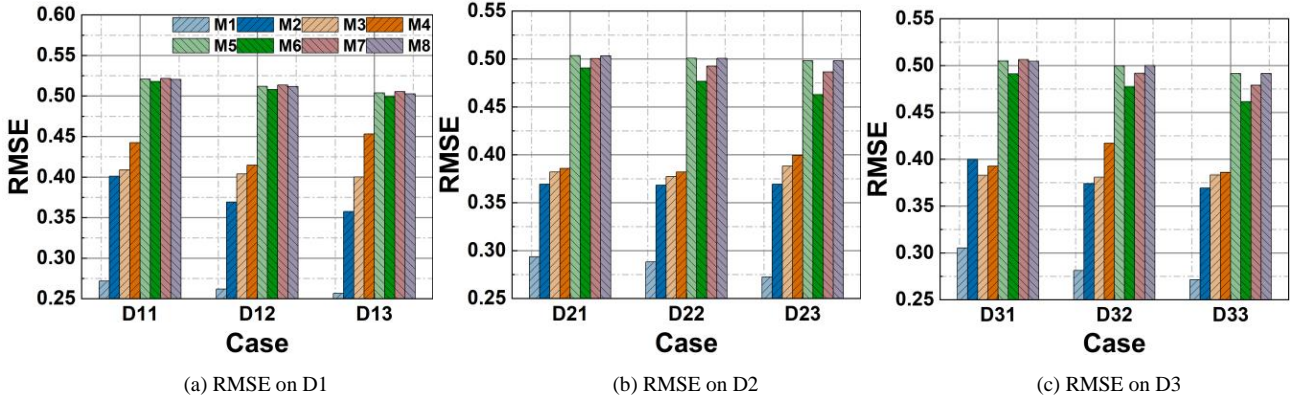

(a) RMSE on D1  (b) RMSE on D2  (c) RMSE on D3

Figure 4. Compared models on RMSE based on different cases.

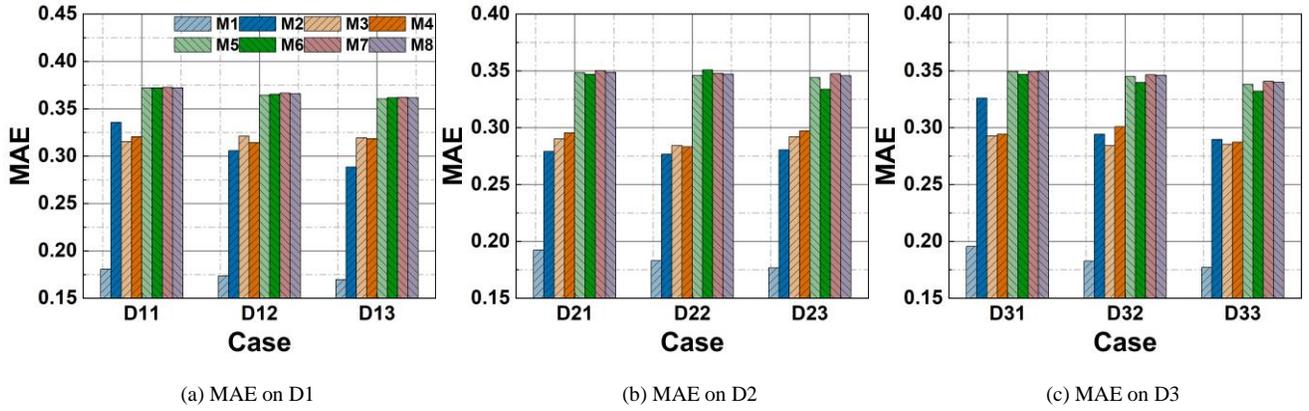

(a) MAE on D1  (b) MAE on D2  (c) MAE on D3

Figure 5. Compared models on MAE based on different cases.

## V. CONCLUSION

This study proposes an EKLF model that effortlessly integrates the principle of an EKF into its learning framework to represent a DWDG accurately from the model-driven perspective. In the future, we plan to adopt an EKF to build a novel dynamic graph neural network [67-75].